\documentclass[conference]{IEEEtran}
\IEEEoverridecommandlockouts
\usepackage{booktabs}
\usepackage{makecell}
\usepackage{tabularray}
\usepackage{cite}
\usepackage{float}
\usepackage{amsmath,amssymb,amsfonts}

\usepackage{array}
\usepackage[longtable]{multirow}
\usepackage{longtable}

\usepackage{graphicx}
\usepackage{textcomp}
\usepackage{xcolor}
\def\BibTeX{{\rm B\kern-.05em{\sc i\kern-.025em b}\kern-.08em
    T\kern-.1667em\lower.7ex\hbox{E}\kern-.125emX}}

\usepackage{algorithm}
\usepackage{algpseudocode}

\begin{document}

\title{Enhancing Transformer with GNN Structural Knowledge via Distillation: A Novel Approach}

\author{\IEEEauthorblockN{ Zhihua Duan}
\IEEEauthorblockA{\textit{Intelligent Cloud Network Monitoring Department} \\
\textit{China Telecom Shanghai Company}\\
\textit{700 Daning Road, Shanghai, 200072}\\
Shanghai,China \\
duanzh.sh@chinatelecom.cn}
\and
\IEEEauthorblockN{ Jialin Wang*}
\IEEEauthorblockA{\textit{Executive Vice President} \\
\textit{Ferret Relationship Intelligence}\\
\textit{Burlingame, CA 94010, USA}\\
https://www.linkedin.com/in/starspacenlp \\
jialinwangspace@gmail.com}
 
}

\maketitle

\begin{abstract}
Integrating the structural inductive biases of Graph Neural Networks (GNNs) with the global contextual modeling capabilities of Transformers represents a pivotal challenge in graph representation learning. While GNNs excel at capturing localized topological patterns through message-passing mechanisms, their inherent limitations in modeling long-range dependencies and parallelizability hinder their deployment in large-scale scenarios. Conversely, Transformers leverage self-attention mechanisms to achieve global receptive fields but struggle to inherit the intrinsic graph structural priors of GNNs. This paper proposes a novel knowledge distillation framework that systematically transfers multiscale structural knowledge from GNN teacher models to Transformer student models, offering a new perspective on addressing the critical challenges in cross-architectural distillation. The framework effectively bridges the architectural gap between GNNs and Transformers through micro-macro distillation losses and multiscale feature alignment. This work establishes a new paradigm for inheriting graph structural biases in Transformer architectures, with broad application prospects. 
\end{abstract}
 

\section{Introduction}
The field of graph representation learning is facing a critical challenge: how to integrate the local structural awareness of Graph Neural Networks (GNNs) with the global contextual modeling advantages of Transformers. Although GNNs excel at capturing neighborhood topological patterns through message-passing mechanisms, their paradigm limits the modeling of long-range dependencies. On the other hand, Transformers can achieve global information interaction through self-attention mechanisms but struggle to inherit the inherent graph structural inductive biases of GNNs. This fundamental contradiction at the architectural level leads to bottlenecks in existing methods for tasks that require simultaneous handling of local structures and global contextual dependencies.  

Current knowledge distillation research primarily focuses on knowledge transfer between homogeneous architectures (e.g., GNN-to-GNN or Transformer-to-Transformer), overlooking the geometric semantic gap in cross-architectural distillation. Traditional methods often rely on simple feature imitation or output distribution alignment strategies, failing to effectively transfer multiscale structural information in graph data. More critically, existing frameworks lack systematic solutions to the following issues: 
\begin{itemize}
    \item How to decompose and transfer the hierarchical structural knowledge implicitly learned by GNNs
    \item How to dynamically balance task-specific objectives and structural preservation objectives
    \item How to ensure the generalization of the distillation process to diverse teacher architectures
\end{itemize}

This paper proposes a  multiscale structural distillation framework, with its main contributions including:
\begin{itemize}
    \item Hierarchical Knowledge Transfer Mechanism: A systematic transfer of GNN structural biases to Transformers is achieved through a dual-perspective distillation strategy at both micro (edge-level distribution alignment) and macro (graph-level topology matching) levels.
    \item Dynamic Optimization Paradigm: An adaptive loss weighting method is introduced to achieve an optimal trade-off between task-driven supervision and structural knowledge preservation, surpassing the limits of traditional static weighting schemes.
\end{itemize}

This work establishes a new paradigm for deploying graph-aware Transformers. Theoretical analysis further reveals that the multiscale alignment mechanism can effectively mitigate the "receptive field mismatch" problem caused by architectural heterogeneity, providing new methodological insights for cross-modal knowledge transfer research.

\section{PRELIMINARY}

% \label{sec:preliminary}

\subsection{Graph Representation}
Let $\mathcal{G} = (\mathcal{V}, \mathcal{E}, \mathbf{X})$ denote a graph with node set $\mathcal{V} = \{v_1,...,v_N\}$, edge set $\mathcal{E} \subseteq \mathcal{V} \times \mathcal{V}$, and node feature matrix $\mathbf{X} \in \mathbb{R}^{N \times d}$. The graph structure is encoded in adjacency matrix $\mathbf{A} \in \{0,1\}^{N \times N}$, where $\mathbf{A}_{ij} = 1$ iff $(v_i, v_j) \in \mathcal{E}$. For node-level tasks, each node $v_i$ has an associated label $y_i \in \mathcal{Y}$.

Message Passing Neural Networks:Modern GNNs follow the message passing paradigm where node representations $\mathbf{h}_v^{(l)}$ at layer $l$ are computed through:

\begin{equation}
\mathbf{m}_v^{(l)} = \text{AGGREGATE}^{(l)}\left(\{\mathbf{h}_u^{(l-1)}: u \in \mathcal{N}(v)\}\right)
\end{equation}

\begin{equation}
\mathbf{h}_v^{(l)} = \text{UPDATE}^{(l)}\left(\mathbf{h}_v^{(l-1)}, \mathbf{m}_v^{(l)}\right)
\end{equation}

where $\mathcal{N}(v)$ denotes the neighborhood of node $v$, AGGREGATE$^{(l)}(\cdot)$ is a permutation-invariant function, and UPDATE$^{(l)}(\cdot)$ combines previous states with aggregated messages.

\subsection{Transformer Architecture}
% \label{subsec:transformer}

Multi-Head Attention Mechanism:
The core component of Transformer is defined through two fundamental equations:

Scaled Dot-Product Attention:
\begin{equation}
\text{Attention}(\mathbf{Q},\mathbf{K},\mathbf{V}) = \text{softmax}\left(\frac{\mathbf{Q}\mathbf{K}^\top}{\sqrt{d_k}}\right)\mathbf{V}
\end{equation}

Multi-Head Projection:
\begin{equation}
\text{MultiHead}(\mathbf{Q},\mathbf{K},\mathbf{V}) = \text{Concat}(\text{head}_1,...,\text{head}_h)\mathbf{W}^O
\end{equation}
where each head $\text{head}_i = \text{Attention}(\mathbf{Q}\mathbf{W}_i^Q, \mathbf{K}\mathbf{W}_i^K, \mathbf{V}\mathbf{W}_i^V)$ with projection matrices $\{\mathbf{W}_i^Q,\mathbf{W}_i^K \in \mathbb{R}^{d_{\text{model}}\times d_k}\}$, $\mathbf{W}_i^V \in \mathbb{R}^{d_{\text{model}}\times d_v}\}$, and $\mathbf{W}^O \in \mathbb{R}^{hd_v\times d_{\text{model}}}\}$.

\section{Related Work}
\subsection{Graph Neural Networks (GNNs)}

The review proposes a new taxonomy that categorizes existing GNNs into four types: recurrent GNNs, convolutional GNNs, graph autoencoders, and spatiotemporal GNNs\cite{Comprehensive}.It systematically discusses various variants of GNNs, such as Graph Convolutional Networks (GCN), Graph Attention Networks (GAT), and Graph Recurrent Networks (GRN), and proposes a general framework for designing GNN models\cite{Graph}.The over-smoothing problem in Graph Neural Networks is systematically analyzed, and a new deep Graph Neural Network model (DAGNN) is proposed, effectively addressing the performance degradation issue in deep network training\cite{Towards}.The GraphPrompt framework integrates pretraining and downstream tasks into a unified task template and employs learnable prompts to bridge the gap between pretraining and downstream task objectives\cite{GraphPrompt}.The process of decoupled Graph Neural Networks (decoupled GNNs) is improved by introducing a scalable attention mechanism to effectively utilize multi-hop information. By regarding label propagation as a special case of decoupled GNNs, a decoupled label technique (DecLT) is proposed to integrate label information\cite{Scalable}. 

\subsection{Graph Knowledge Distillation}
Recently, knowledge distillation has been proven to be effective in the field of graph learning.A new knowledge distillation framework named MuGSI is proposed for graph classification. It addresses two main challenges in the application of existing knowledge distillation frameworks to graph classification, namely the inherent sparsity of learning signals and the expressive limitations of student MLPs, through multi-granular structural information \cite{MuGSI}. A full-frequency band GNN-to-MLP distillation framework is proposed. It extracts low-frequency and high-frequency knowledge from GNNs and injects it into MLPs, solving the problem of information loss in traditional GNN-to-MLP distillation \cite{GNN-to-MLP}. By measuring the invariance of the information entropy of GNNs to noise perturbations, the reliability of knowledge points is quantified, and a method of Knowledge Heuristic Reliable Distillation (KRD) is proposed. This method can identify and utilize reliable knowledge points for the training of student MLPs \cite{Quantifying}. A multi-task self-distillation framework is proposed. By introducing self-supervised learning and self-distillation techniques, it addresses the mismatch problem of graph convolutional networks in graph-based semi-supervised learning from the graph structure side and the label side respectively \cite{Multi-task}. A novel adaptive knowledge distillation framework BGNN is proposed. By sequentially transferring knowledge from multiple GNNs to a student GNN, and using an adaptive temperature module and a weight enhancement module to guide the student to learn effectively, it achieves improvements in both node classification and graph classification tasks \cite{Boosting}. 

Unlike prior works, this paper proposes a novel knowledge distillation framework that systematically transfers the multiscale structural knowledge of Graph Neural Networks (GNNs) to Transformer student models. By bridging the architectural gap through micro-macro distillation losses and multiscale feature alignment, it establishes a new approach for distilling graph structural information into Transformer architectures.

\section{Methods}
Figure 1 illustrates the design of a knowledge - distillation system from a Graph Neural Network (Teacher GNN) to a Transformer (Student Transformer). Through a hierarchical knowledge - distillation mechanism, it enables cross - architectural knowledge transfer between Graph Neural Networks (GNNs) and Transformers. 

\begin{figure*}[htbp]
  \centering
  \includegraphics[width=\linewidth]{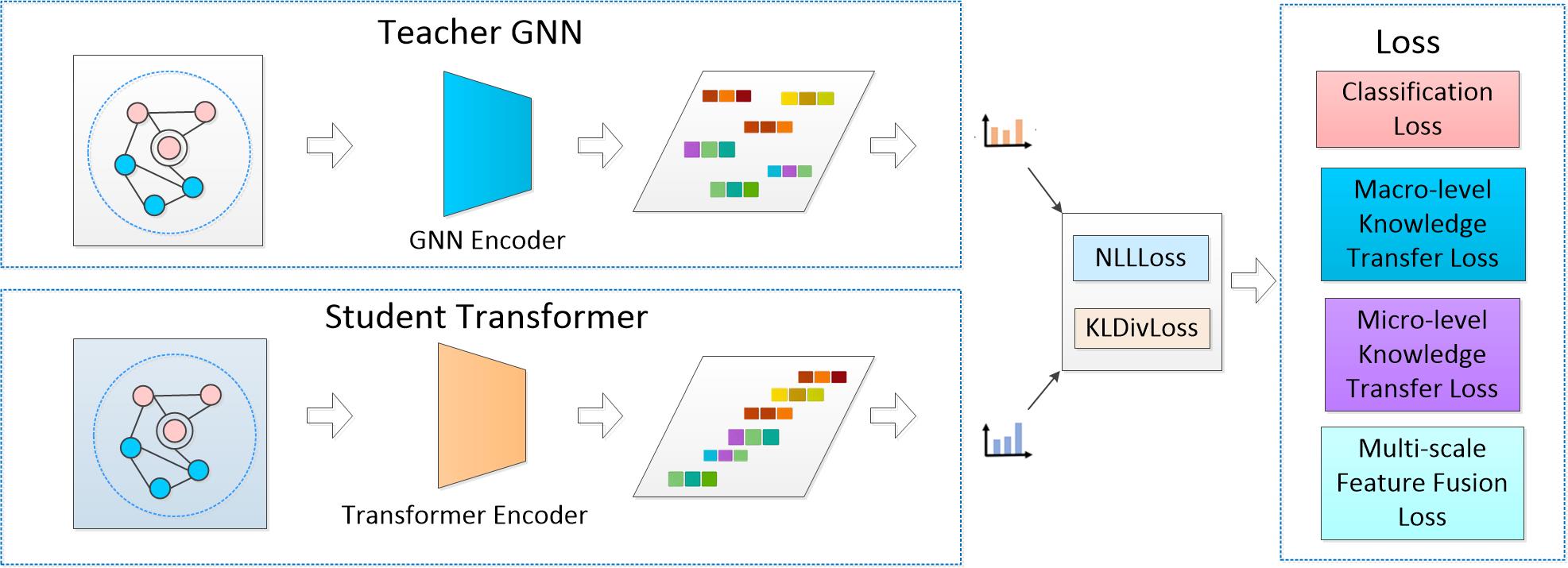}
  \caption{Knowledge Distillation from GNNs to Transformers .} 
\end{figure*}

\subsection{Teacher Graph Neural Network (Teacher GNN)}
The teacher model employs a graph neural network-based encoding architecture for structural data processing. Through hierarchical neighborhood aggregation operations, this framework performs graph representation learning via node-wise feature extraction and latent space embedding, ultimately generating a node embedding matrix that encodes both local graph topology and global structural semantics.

\subsection{Student Transformer (Student Transformer)}
The student architecture receives the teacher's latent representations as input and processes them through a transformer encoder module. This module leverages multi-head self-attention mechanisms to enable global context-aware feature transformation, thereby capturing long-range dependencies and complex graph interactions in the generated token-level embeddings.

Through the systematic transfer of GNN structural biases to Transformers, the geometric - semantic gap in cross - architectural distillation is bridged. It supports stable distillation of heterogeneous teacher architectures (GCN/GAT/GraphSAGE).

\section{Loss Function Formulation}
Our distillation framework employs four synergistic loss components to transfer structural knowledge from the teacher GNN to the student Transformer:

\subsection{Classification Loss}
The task-specific supervision using Negative Log-Likelihood (NLL) loss:

% \begin{equation}
% \mathcal{L}_{\text{cls}} = -\frac{1}{|\mathcal{V}_L|}\sum_{v \in \mathcal{V}_L} \sum_{c=1}^C y_v^{(c)} \log p_S^{(c)}(v)
% \end{equation}

\begin{equation}
\mathcal{L}_{\text{cls}} = -\frac{1}{|\mathcal{V}_L|} \sum_{v \in \mathcal{V}_L} \mathbf{y}_v^\top \log\left(\text{softmax}(\mathbf{z}_v^S)\right)
\end{equation}

where $\mathcal{V}_L$ denotes labeled nodes, $y_v$ the label vector, and $p_S(v) = \text{softmax}(\mathbf{z}_v^S)$ the student's prediction distribution.

% The classification loss, denoted by $\mathcal{L}_{\text{cls}}$, measures the discrepancy between the predicted labels and the ground-truth labels using the Negative Log Likelihood Loss (NLLLoss). It is defined as:
% \begin{equation}
% \mathcal{L}_{\text{cls}} = -\sum_{i \in \mathcal{I}_l} y_i \log(\hat{y}_i),
% \end{equation}
% where $\mathcal{I}_l$ represents the indices of labeled data, $y_i$ is the ground-truth label for the $i$-th sample, and $\hat{y}_i$ is the predicted probability of the $i$-th sample belonging to the corresponding class.

\subsection{Micro-Structure Distillation}
Edge-wise distribution alignment via Kullback-Leibler (KL) divergence:

\begin{equation}
\mathcal{L}_{\text{micro}} = \frac{1}{|\mathcal{E}|}\sum_{(u,v) \in \mathcal{E}} D_{\text{KL}}\left(p_T(v) \parallel p_S(u)\right)
\end{equation}

where $p_T(v) = \text{log-softmax}(\mathbf{z}_v^T/\tau)$ represents the teacher's distribution at head node $v$, and $p_S(u) = \text{softmax}(\mathbf{z}_u^S)$ the student's distribution at tail node $u$.

\subsection{Macro-Structure Distillation}
Graph-level distribution matching with temperature scaling:

\begin{equation}
\mathcal{L}_{\text{macro}} = D_{\text{KL}}\left(q_S^{\text{high}} \parallel q_T^{\text{high}}\right)
\end{equation}

where the high-level edge distributions are computed as:

\begin{align}
q_S^{\text{high}} &= \text{log-softmax}\left(\frac{||\mathbf{z}_u^S - \mathbf{z}_v^S||_1}{\tau}\right) \\
q_T^{\text{high}} &= \text{log-softmax}\left(\frac{||\mathbf{z}_u^T - \mathbf{z}_v^T||_1}{\tau}\right)
\end{align}

for each edge $(u,v) \in \mathcal{E}$.

\subsection{Multi-Scale Feature Consistency}
Edge feature consistency across scales:

\begin{equation}
\mathcal{L}_{\text{multi}} = \frac{1}{K}\sum_{k=1}^K D_{\text{KL}}\left(\mathbf{m}_k \parallel \bar{\mathbf{m}}\right)
\end{equation}

where $\mathbf{m}_k$ denotes edge features at scale $k$ and $\bar{\mathbf{m}} = \frac{1}{K}\sum_{k=1}^K \mathbf{m}_k$ the multi-scale average.

\subsection{Total Objective}
The final optimization target combines all components:

\begin{equation}
\mathcal{L}_{\text{total}} = \lambda \mathcal{L}_{\text{cls}} + (1-\lambda)(\mathcal{L}_{\text{micro}} + \mathcal{L}_{\text{macro}} + \mathcal{L}_{\text{multi}})
\end{equation}

with $\lambda \in [0,1]$ controlling the supervision balance.

\section{Experimental Design}

\subsection{Dataset}

The Citeseer dataset is an academic literature citation network dataset created by researchers at Cornell University. It contains 3,327 scientific research papers, which are divided into 6 fields. Each paper is represented by a feature vector, and the edges between papers represent citation relationships. This dataset is often used for node classification tasks in Graph Neural Networks (GNNs), academic network analysis, and text classification research. It is an important benchmark dataset for evaluating the performance of graph algorithms.

\subsection{Experimental Results}

As shown in Table 1, this study systematically evaluated the knowledge distillation framework based on graph neural networks on the Citeseer benchmark dataset. The key components of the experimental design are: (1) The teacher models adopt different message passing mechanisms (GCN, GraphSAGE, GAT); (2) The student model is a Transformer architecture based on multi-head self-attention.

\begin{table}
\begin{center}
\caption{Performance of Different GNN Teacher Models on Citeseer}
\end{center}
\centering
\renewcommand{\arraystretch}{1.5} % 增加行高，倍数可按需调整
\begin{tabular}{>{\hspace{0pt}}m{0.337\linewidth}>{\hspace{0pt}}m{0.26\linewidth}>{\hspace{0pt}}m{0.24\linewidth}} 
\hline
Teacher GNNs                                          & Method       & Citeseer      \\ 
\hline
\multirow{4}{0.337\linewidth}{\hspace{0pt}GCN~~}      & Vanilla GCN$^{\dag}$  & 71.6  ± 0.4   \\
                                                      & Vanilla MLP$^{\dag}$  & 60.7  ± 0.5   \\
                                                      & GLNN$^{\dag}$         & 72.7  ± 0.4   \\ 
\cline{2-3}
                                                      & Transformer  & 74.5  ± 0.4   \\ 
\hline
\multirow{4}{0.337\linewidth}{\hspace{0pt}GraphSAGE~} & Vanilla SAGE$^{\dag}$ & 70.9  ± 0.6   \\
                                                      & Vanilla MLP$^{\dag}$  & 60.7  ± 0.5   \\
                                                      & GLNN$^{\dag}$         & 70.5  ± 0.5   \\ 
\cline{2-3}
                                                      & Transformer  & 67.5  ± 0.4   \\ 
\hline
\multirow{4}{0.337\linewidth}{\hspace{0pt}GAT~~}      & Vanilla GAT$^{\dag}$  & 71.2  ± 0.5   \\
                                                      & Vanilla MLP$^{\dag}$  & 60.7  ± 0.5   \\
                                                      & GLNN$^{\dag}$        & 70.6  ± 0.8   \\ 
\cline{2-3}
                                                      & Transformer  & 72.56  ± 0.5  \\
\hline
\end{tabular}
\footnotesize{ Scores with $^{\dag}$ were obtained from the paper\cite{GNN-to-MLP}}\\
\end{table}

1. \text{Analysis of the GCN Teacher Model}:
On the Citeseer dataset, the standard GCN architecture (Vanilla GCN) achieved a classification accuracy of 71.6 $\pm$ 0.4, while the multi-layer perceptron baseline model (MLP) only obtained 60.7 $\pm$ 0.5. The GNN-Transformer framework achieved an accuracy of 74.5 $\pm$ 0.4, which is better than the GLNN method (72.7 $\pm$ 0.4), verifying the effectiveness of cross-architecture feature distillation.

2. \text{Analysis of the GraphSAGE Teacher Model}:
When GraphSAGE is used as the teacher, the Vanilla MLP shows information degradation on the Citeseer dataset (60.7 $\pm$ 0.5). The GNN-Transformer successfully transferred the neighborhood expansion pattern of the teacher model through the hierarchical attention mechanism, and it is relatively stable on the Citeseer dataset, achieving 67.5 $\pm$ 0.4.

3. \text{Analysis of the GAT Teacher Model}:
The Vanilla GAT model achieved a baseline performance of 71.2 $\pm$ 0.5 on Citeseer. The GNN-Transformer framework, through designing the attention head mechanism, enables the student Transformer to accurately capture the pattern of the teacher model, and finally achieves a performance of 72.56 $\pm$ 0.5.

\section{Discussion}
Although this research has made progress in transferring the structural knowledge of Graph Neural Networks (GNNs) to Transformers, there are still some areas for improvement and optimization.
In terms of model training efficiency, the current distillation framework involves multiple complex components when calculating the loss function, such as micro - structure distillation, macro - structure distillation, and multi - scale feature loss. This leads to a large amount of computation in the training process. More efficient calculation methods can be explored to optimize the calculation process of the loss function.
From the perspective of model generalization, although good results have been achieved on the Citeseer dataset, graph data in different fields and tasks vary greatly in structure, scale, and features. Therefore, the generalization of the model in other datasets or practical application scenarios needs to be further verified and improved.

In future cross-task and cross-domain research, as application scenarios diversify, the deep integration of GNNs and LLMs is expected to be key for solving complex graph data problems. For instance, by combining GNNs' graph  structuring and DeepSeek's language understanding, graph data's semantic and structural features can be processed more efficiently, with broad application prospects. 

\section{Conclusion}
This paper proposes a knowledge distillation framework that transfers the structural knowledge of Graph Neural Networks (GNNs) into Transformers. By integrating classification loss, micro-structure distillation, macro-structure distillation, and multi-scale distillation into a unified objective function, it can effectively capture both local and global structural information from GNNs and transfer it to Transformers. The experimental results on the Citeseer dataset demonstrate the effectiveness of the proposed method. This research provides a new idea for the combination of Graph Neural Networks and Transformers, and offers valuable implementations for the design of cross-architectural knowledge distillation mechanisms.  
 
\section{Acknowledgement} 
This study is inspired by GraphPrompt  and FF-G2M projects.

% \bibliographystyle{unsrt} 
% \bibliography{GNN} 

\end{document}